\def\year{2020}\relax
\documentclass[letterpaper]{article} 
\usepackage{aaai20}  
\usepackage{times}  
\usepackage{helvet} 
\usepackage{courier}  
\usepackage[hyphens]{url}  
\usepackage{graphicx} 
\usepackage{graphicx}  
\usepackage{xcolor} 
\usepackage{amsmath} 
\usepackage{amssymb} 
\usepackage{soul} 
\usepackage{booktabs} 
\usepackage{subfig} 
\title{Machine Learning Approaches in Agile Manufacturing with Recycled Materials for Sustainability}

\author{Aparna S. Varde\\
Department of Computer Science \\
 Clean Energy \& Sustainability Analytics Center\\
 Montclair State University, NJ, USA\\
vardea@montclair.edu,  ORCID ID: 0000-0002-3170-2510\\
 \And
 Jianyu Liang\\
 Mechanical and Materials Engineering\\
 Worcester Polytechnic Institute\\
 Worcester, MA, USA \\
 jianyul@wpi.edu
 }

\begin{document}
\newcommand{\ourmodel}[1]{\textsc{CSK-SNIFFER}}
\newcommand{\yolo}[1]{\textsc{YOLO}}
 
\definecolor{Red}{rgb}{1,0,0}
\definecolor{Green}{rgb}{0,0.7,0}
\definecolor{Blue}{rgb}{0,0,1}
\definecolor{Red}{rgb}{0.6,0,0}
\definecolor{Orange}{rgb}{1,0.5,0}
\newcommand{\niket}[1]{\textcolor{Green}{[#1 \textsc{--Niket}]}}
\newcommand{\anurag}[1]{\textcolor{Blue}{[#1 \textsc{--Anurag}]}}
\newcommand{\aparna}[1]{\textcolor{Red}{[#1 \textsc{--Aparna}]}}
\newcommand{\reviewed}[1]{\textcolor{Orange}{[#1]}}

\makeatletter
\newcommand*\bigcdot{\mathpalette\bigcdot@{.5}}
\newcommand*\bigcdot@[2]{\mathbin{\vcenter{\hbox{\scalebox{#2}{$\m@th#1\bullet$}}}}}
\makeatother

\renewcommand{\v}[1]{$\mathbf{#1}$}
\newcommand{\vect}[1]{\mathbf{#1}}

\newcommand{\bluebox}[1]{\colorbox{blue!10}{#1}}
\newcommand{\redbox}[1]{\colorbox{red!10}{#1}}
\newcommand{\purplebox}[1]{\colorbox{purple!10}{#1}}



\def\DG{{\mathcal{G}}}

\newtheorem{theorem}{Definition}[section]

\newcommand{\statechange}[1]{\texttt{\textit{#1}}}
\newcommand{\entity}[1]{\texttt{#1}}
\newcommand{\strikethrough}[1]{\st{#1}}

\newcommand{\namecite}[1]{\citeauthor{#1}~\shortcite{#1}}
\newcommand{\com}[1]{}
\newcommand{\myparagraph}[1]{\vspace{1mm} \noindent {\bf #1: }}
\newcommand{\bfit}[1]{\textbf{\textit{#1}}}
\newcommand{\eat}[1]{}
\mathchardef\mhyphen="2D
\newenvironment{ite}{                     
     \parskip 0cm \begin{itemize} \parskip 0cm \parsep 0cm \itemsep 0cm \topsep 0cm}{
        \end{itemize}} 
\newenvironment{enu}{                   
     \parskip 0cm \begin{list}{}{\parsep 0cm \itemsep 0cm \topsep 0cm}}{
      \end{list}} 
\newenvironment{des}{                 
     \parskip 0cm \begin{list}{}{\parsep 0cm \itemsep 0cm \topsep 0cm}}{
      \end{list}} 
\newenvironment{myenumerate}{                   
     \parskip 0cm \begin{enumerate}{\parsep 0cm \itemsep 0cm \topsep 0cm}}{
        \end{enumerate}} 
\newenvironment{myitemize}{                     
     \parskip 0cm \begin{itemize}{\parsep 0cm \itemsep 0cm \topsep 0cm}}{
        \end{itemize}} 
\newcommand{\ourdataexpansion}{``What-If Question Answering''}
\newenvironment{myquote}{                   
  \parskip 0mm \begin{quoting}[vskip=0mm,leftmargin=2mm]}{
\end{quoting}}
\newcommand{\red}[1]{\textcolor{red}{#1}}
\newcommand{\blue}[1]{\textcolor{blue}{#1}}
\newcommand{\green}[1]{\textcolor{green}{#1}}
\newenvironment{mycentering}
 {\parskip=0pt\par\nopagebreak\centering}
 {\par\noindent\ignorespacesafterend}

    

\maketitle

\begin{abstract}
It is important to develop sustainable processes in materials science and manufacturing that are environmentally friendly. AI can play a significant role in decision support here as evident from our earlier research leading to tools developed using our proposed machine learning based approaches. Such tools served the purpose of computational estimation and expert systems. This research addresses environmental sustainability in materials science via decision support in agile manufacturing using recycled and reclaimed materials. It is a safe and responsible way to turn a specific waste stream to value-added products. We propose to use data-driven methods in AI by applying machine learning models for predictive analysis to guide decision support in manufacturing. This includes harnessing artificial neural networks to study parameters affecting heat treatment of materials and impacts on their properties; deep learning via advances such as convolutional neural networks to explore grain size detection; and other classifiers such as Random Forests to analyze phrase fraction detection. Results with all these methods seem promising to embark on further work, e.g. ANN yields accuracy around 90\% for predicting micro-structure development as per quench tempering, a heat treatment process. Future work entails several challenges: investigating various computer vision models (VGG, ResNet etc.) to find optimal accuracy, efficiency and robustness adequate for sustainable processes; creating domain-specific tools using machine learning for decision support in agile manufacturing; and assessing impacts on sustainability with metrics incorporating the appropriate use of recycled materials as well as the effectiveness of developed products. Our work makes impacts on green technology for smart manufacturing, and is motivated by related work in the highly interesting realm of AI for materials science.
\end{abstract}

\section{Introduction}
The role of AI in the overall realm of sustainability is critical in recent times. This is because we truly need to live in a green, clean and sustainable manner in order to save the planet, wherein machine learning techniques can help to make adequate predictions for decision support in scientific processes. This is pertinent to various scientific domains as evident from a plethora of studies in the literature \cite{butler2018machine}, \cite{Kothoff2022}, \cite{nishant2020artificial}, \cite{ai2011large}, \cite{hutter2019automated}, \cite{rajasekar2009application}, \cite{liu2022towards}, \cite{tsolakis2021towards}, \cite{Weikum2021}, \cite{Zaki2010}, \cite{Zadeh2015}. In this paper, we address the domain of Materials Science \& Manufacturing, claiming that sustainable processes are imperative. 

More specifically, our problem is defined as follows. There is a necessity to develop agile manufacturing techniques using recycled \& reclaimed metals. It is a safe \& responsible method of turning a specific waste stream to value-added products, in line with environmental sustainability. Note that agile manufacturing is ``a manufacturing methodology that places an extremely strong focus on rapid response to the customer - turning speed and agility into a key competitive advantage'' \cite{AgileMfg}. Therefore, agile manufacturing is a recommended approach for acquiring a ``competitive advantage in the fast-moving marketplace'' of the modern day. We investigate the contributions of AI here, for developing tools based on machine learning to advance sustainable agile manufacturing. 

\section{Earlier Research: AutoDomainMine}

\begin{figure}
    \centering
    {\frame{\includegraphics[width=1.0\columnwidth]{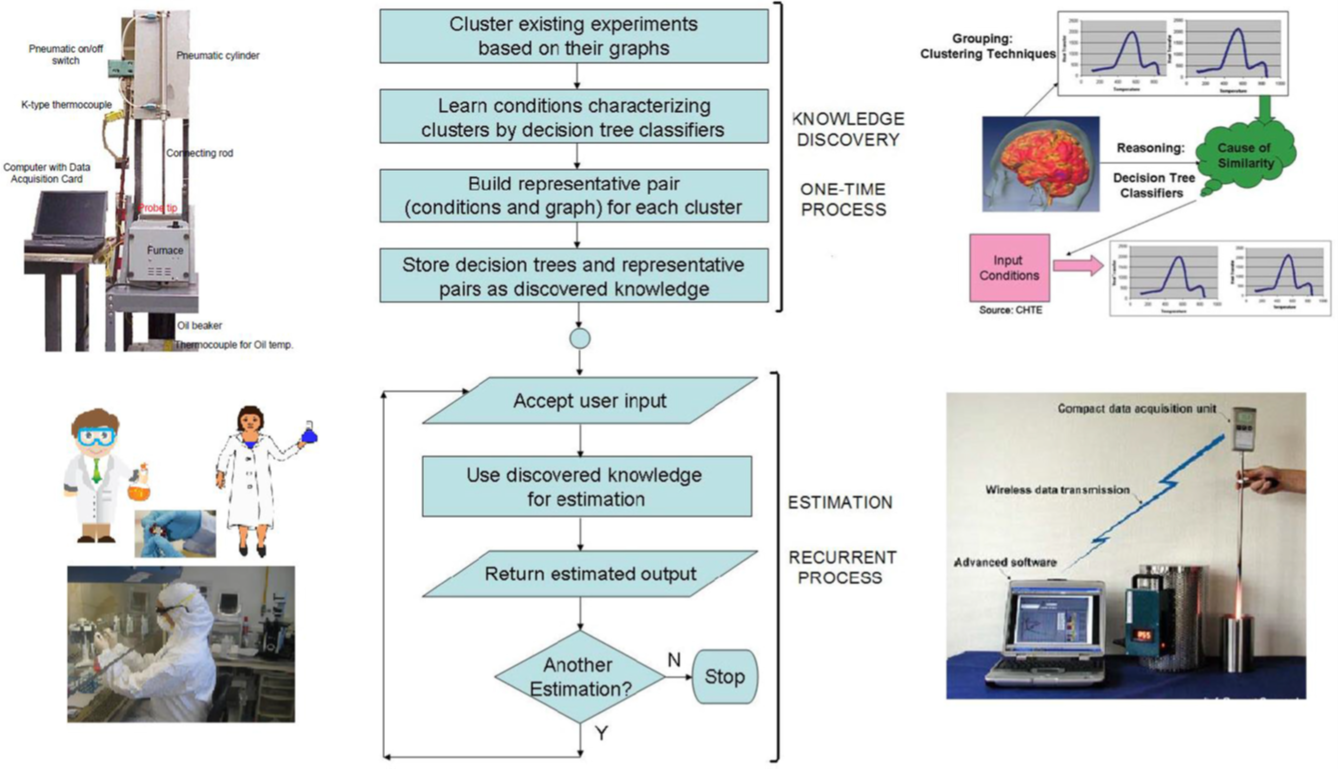}}}
    \caption{AutoDomainMine for computational estimation}
    \label{ADM}
\end{figure}

In our earlier research overlapping AI and Materials Science, we addressed the problem of developing a computational estimation approach such that: (1) given some or all of the input conditions of an experiment, it should display the most likely resulting graph; (2) given desired graph in an experiment or ranges describing its features, it should determine the most appropriate input conditions to achieve it. Our proposed solution entailed an approach called AutoDomainMine  to automate a typical learning strategy of scientists via an integrated framework of data mining techniques (see Figure \ref{ADM}) \cite{Varde2006AAAI}. This approach obtained desired levels of accuracy acceptable for targeted applications. The estimation needed far less time than lab experiments saving significant time \& resources. Domain knowledge was incorporated within AutoDomainMine \& its sub-processes to make them more meaningful for computational estimation. The resulting tool ``AutoDomainMine'' was found extremely useful \cite{Varde2007SIGMOD}. A journal article was recently authored based on its test-of-the-time success story due to numerous citations in CS \& Materials Science, as well as practical usefulness \cite{varde2022computational}. 

Motivated by AutoDomainMine and other such success stories harnessing AI techniques in various scientific domains \& applications \cite{varde2007learning}, \cite{pawlish2010decision}, \cite{varde2009challenging}, \cite{puri2018smart}, \cite{basavaraju2017supervised}, \cite{du2020public}, \cite{pathak2019ubiquitous} we propose AI-related approaches for sustainable manufacturing in our current research as described next.  

\section{Proposed Approaches}
\begin{figure}
    \centering
    {\frame{\includegraphics[width=1.0\columnwidth]{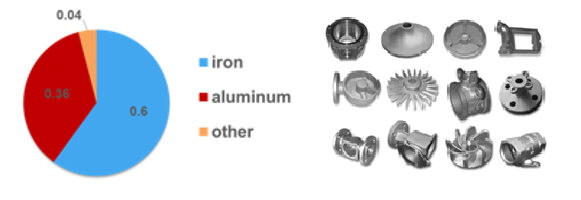}}}
    \caption{Metal waste and its breakdown}
    \label{MetalWaste}
\end{figure}

\begin{figure}
    \centering
    {\frame{\includegraphics[width=1.0\columnwidth]{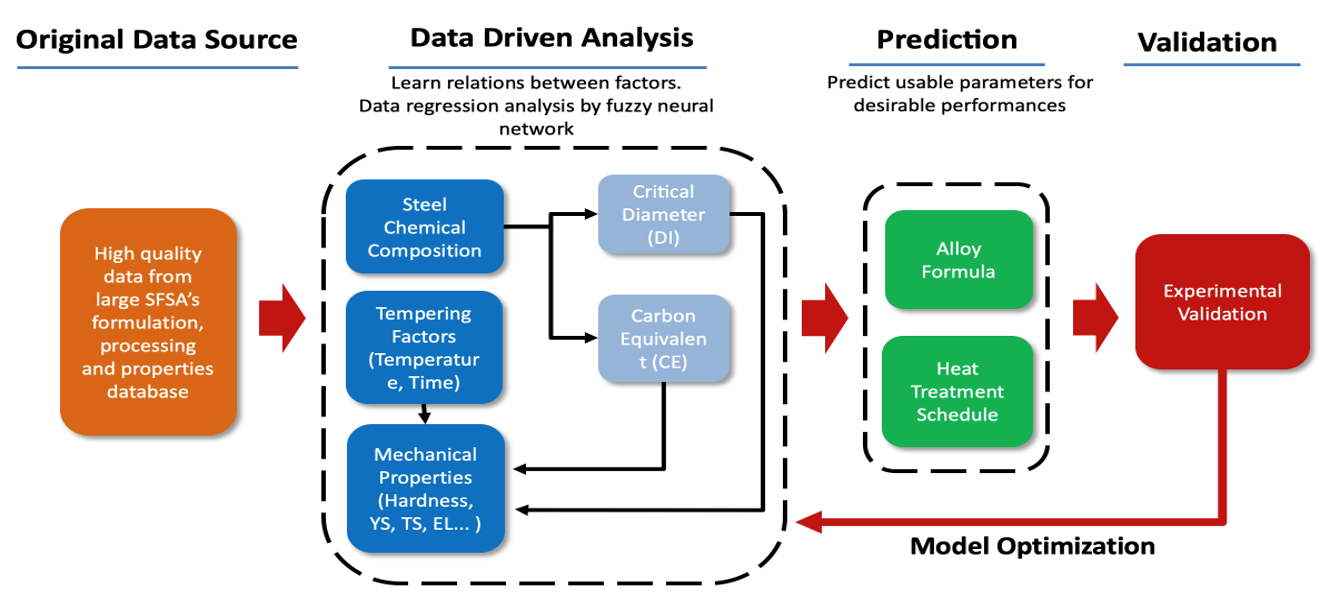}}}
    \caption{Procedure for analysis}
    \label{Analysis}
\end{figure}

\begin{figure}
    \centering
    {\frame{\includegraphics[width=1.0\columnwidth]{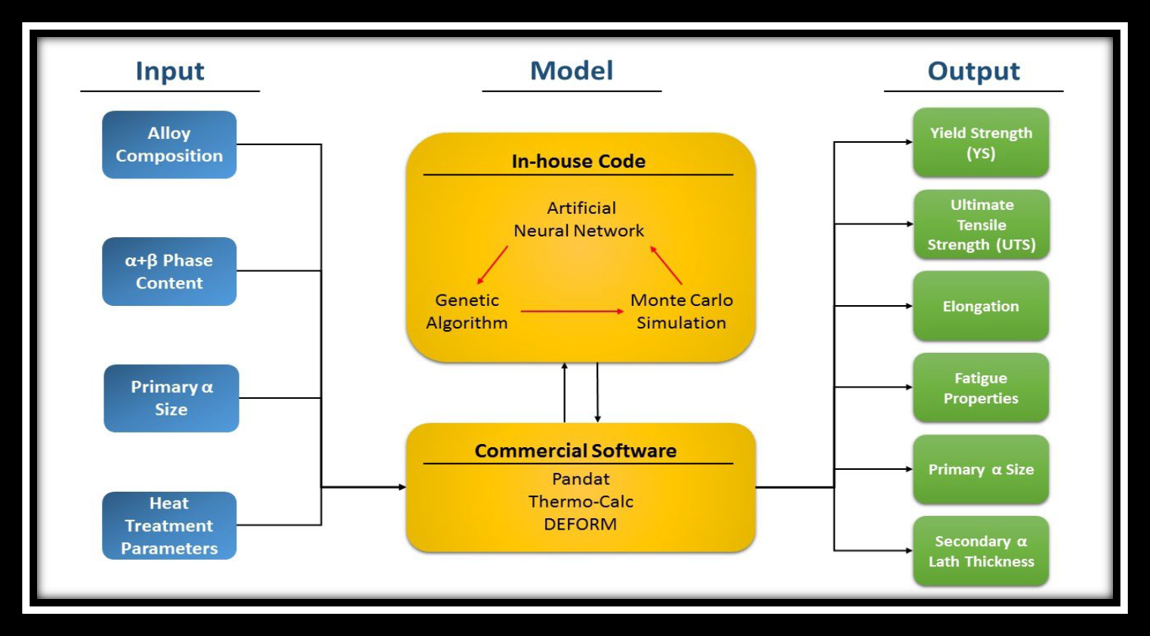}}}
    \caption{Modeling with machine learning techniques}
    \label{ANN-ML}
\end{figure}

\begin{figure}
    \centering
    {\frame{\includegraphics[width=1.0\columnwidth]{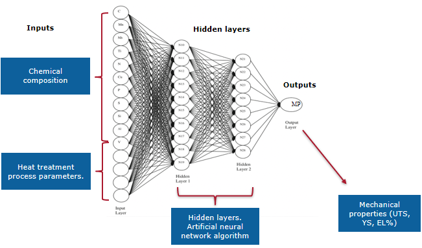}}}
    \caption{ANN: predict mechanical properties of steel alloy}
    \label{ANN-details}
\end{figure}

Our work spans developing data-driven methods and tools to enable rationally designed production. In this respect, we have recently developed machine learning based tools to predict the mechanical properties of steel alloys based on compositions, heat treatments and microstructures \cite{Wang2021MST}. We have also contributed to adapting machine learning related models for predicting the influence of heat treatments on the tensile properties of Ti6Al4V parts prepared by selective laser melting \cite{Wang2022CHTE}. Note that Ti6Al4V is an ``alpha-beta titanium alloy with a high specific strength and excellent corrosion resistance'' \cite{WikiTi}. It among the popular titanium alloys useful in situations where ``low density and excellent corrosion resistance'' are important, e.g. aerospace applications as well as bio-mechanical devices such as dental implants, hip \& knee joints etc. Some of our work in this context entails estimating the tensile properties of annealed Ti6Al4V parts manufactured via selective laser melting \cite{Yang2022AIEDAM}. This research makes impacts on sustainable manufacturing, 

In line with such work, we study the recommended breakdown of metal waste recipe at forward operating bases, as depicted in Figure \ref{MetalWaste}. We aim to progress from waste metal to weapons, thus conducting additive manufacturing enabled agile manufacturing \cite{Liang2022}. We propose a data-driven analysis method, an overview of which is illustrated in Figure \ref{Analysis}. Modeling to optimize heat treatment parameters occurs as in Figure \ref{ANN-ML}. ANNs (Artificial Neural Networks) are deployed for this purpose. Microstructure prediction using a suitable ANN topology is synopsized in Figure \ref{ANN-details}. 

\begin{figure}
    \centering
    {\frame{\includegraphics[width=0.8\columnwidth]{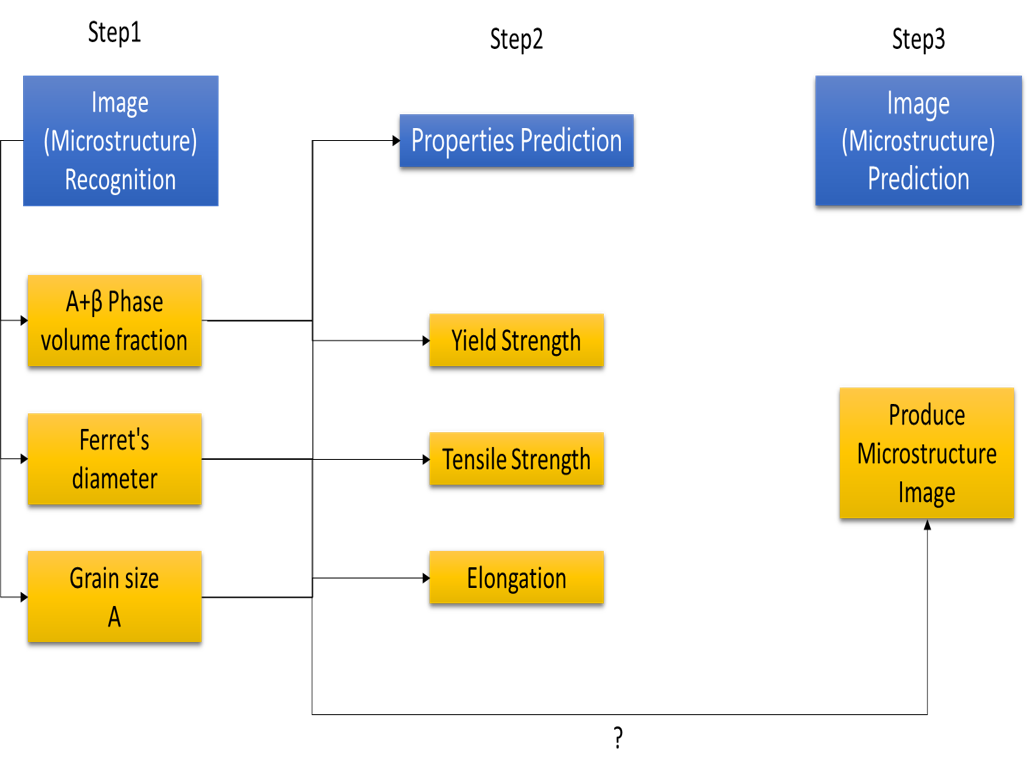}}}
        \caption{Phrase fraction detection by Random Forests}
    \label{RF}
\end{figure}

\begin{figure}
    \centering
    {\frame{\includegraphics[width=0.6\columnwidth]{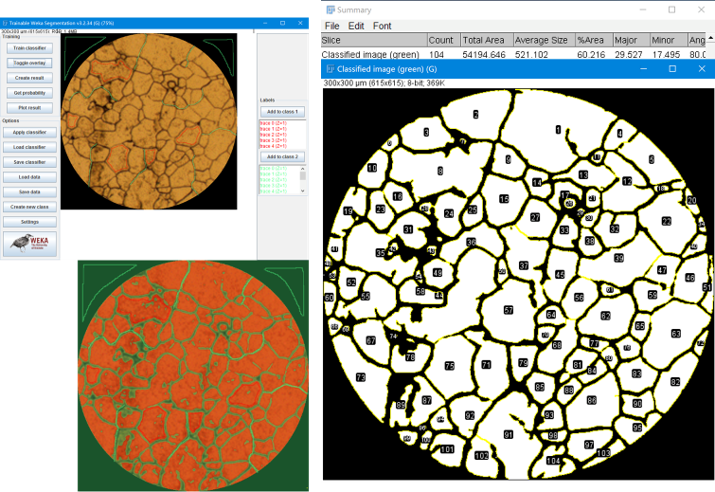}}}
    \caption{Grain-size detection by CNN}
    \label{CNN}
\end{figure}

We conduct further microstructure analysis by adapting deep learning and other classifiers in machine learning. Deep learning occurs with CNNs (Convolutional Neural Networks) to augment the accuracy for learning at a finer level of granularity. Additionally, we deploy RF (Random Forests) as classifiers to enhance interpretability and explainability. Figure \ref {RF} portrays an example of phrase fraction detection by utilizing the RF classifier. Grain-size detection by harnessing CNN is exemplified in Figure \ref {CNN}. 

Likewise, much analysis occurs using various machine learning paradigms in order to guide manufacturing processes for agile as well as sustainable technologies. Machine learning models help to estimate certain tendencies in advance, predict the occurrence of specific phenomena, and thereby assist in decision-making. For instance, if scientists can gauge that a particular combination of input conditions leads to a resulting microstructure in terms of grain-size and other parameters based on discovering knowledge from CNN outputs, they can use that knowledge to make decisions about selecting the corresponding input conditions in real industrial processes and hence advise the concerned working professionals. Thus, in order to conduct manufacturing in an agile sustainable manner, scientists and industrial professionals can plan accordingly by conserving time and resources, using recycled materials, waste products etc. so as to obtain the desired results efficiently. Since the large amounts of data obtained from previously conducted laboratory experiments is used to train the machine learning models rigorously, these models can provide adequate predictive analysis for decision support in real industrial processes. Hence, machine learning can be used to guide agile sustainable manufacturing by acquiring knowledge from existing data and using that to support further decision-making, enhancing accuracy, efficiency as well as clean production. 

\section{Experimental Results}
\begin{figure}
    \centering
    {\frame{\includegraphics[width=1.0\columnwidth]{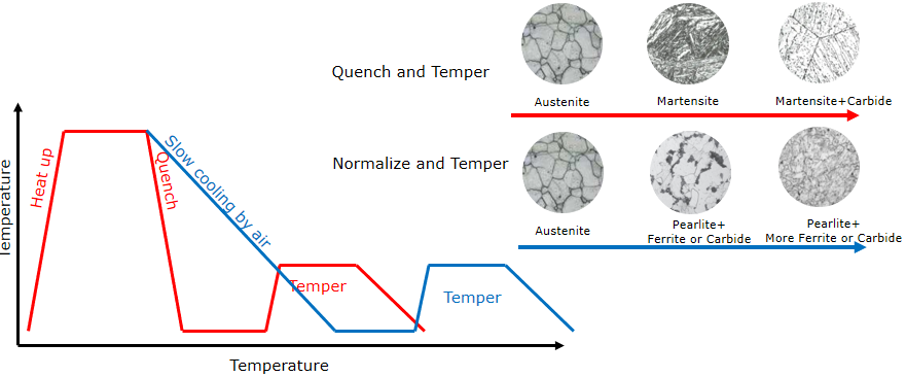}}}
    \caption{Quench \& temper, Normalize \& temper}
    \label{QTNT}
\end{figure}

We present a summary of our exhaustive experimentation in connection with machine learning for agile sustainable manufacturing. 
Laboratory studies on microstructure development in Quench \& Temper (QT) as well as Normalize \& Temper (NT) are used as the basis for learning in ANN experiments. Quenching is a process entailing the rapid cooling of a material in a liquid or gas medium to obtain desired mechanical properties. Tempering, on the other hand, heats the material to a high temperature prior to cooling it via different media, thereby contributing towards decreasing its internal stresses and minimizing its brittleness \cite{EatonSteel}. Normalizing is a related process wherein the steel is heated to a high temperature and thereafter subjected to slow cooling down to room temperature. This heating and slow cooling alters its microstructure, decreases its hardness and raises its ductility \cite{Thermex}. Figure \ref{QTNT} presents examples of our experiments pertaining to these processes. 

\begin{figure}
    \centering
    {\frame{\includegraphics[width=0.6\columnwidth]{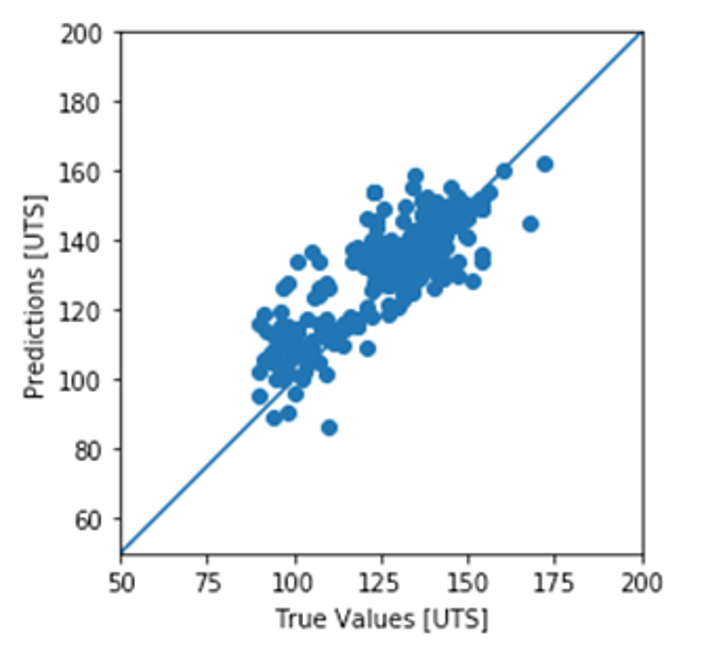}}}
    \caption{ANN modeling results: True values vs. prediction (by chemical composition only)}
    \label{ANN-Plot}
\end{figure}

\begin{figure}
    \centering
    {\frame{\includegraphics[width=0.6\columnwidth]{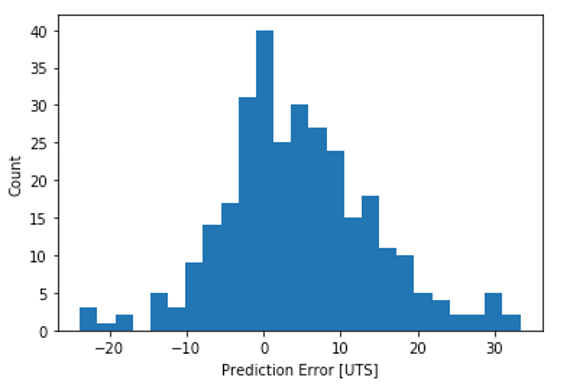}}}
    \caption{ANN modeling results: Count vs. prediction error - Std. error of estimate is 9.81 ksi (kilo pounds per sq. inch}
    \label{ANN-Estimate}
\end{figure}

The ANN prediction results, based on using training data from these processes, are synopsized in Figures \ref{ANN-Plot} and \ref{ANN-Estimate}. More specifically, Figure \ref{ANN-Plot} plots the true values versus the predicted values, considering chemical composition only, as per the ANN experiments on Ultimate Tensile Strength (UTS) for Quench \& Temper (QT). On a related note, Figure \ref{ANN-Estimate} presents a chart of the count versus prediction error in the ANN experiments on UTS for QT. It is observed that the standard error of estimate is 9.81 ksi (kilo pounds per square inch) in this example. Likewise, numerous experiments are conducted using machine learning paradigms such as ANN. Domain experts infer that the results obtained from such experiments are accurate and efficient for predictive analysis. Hence, such studies can be useful to guide decision-making in the corresponding real industrial processes to aid agile sustainable manufacturing. For instance, in some earlier work, we developed an expert system called QuenchMiner \cite{varde2003quench} in the context of heat treatment processes, analyzing tendencies such as distortion, part hardness in quenching etc. thereby guiding decision support in heat treatment analogous to a real human expert. Likewise, results from our experiments here can pave the way for building tools to aid agile sustainable manufacturing. 

\section{Related Work}
There is much work overlapping AI and Materials Science as well as AI and environmental sustainability. Optimization of laser-induced graphene production is proposed in a recent study \cite{Kothoff2022} by generating a series of unique datasets for machine learning with challenges related to behavioral modeling, knowledge transfer and production parameters. The role of data mining and machine learning methods such as decision tree classifiers and case-based reasoning is explored with respect to the greening of data centers \cite{pawlish2010decision}. ANN based solutions are investigated in a recent PhD dissertation \cite{YangZ} within Mechanical Engineering, in the heat treatment of materials.  

On a related note, since there are images involved in many scientific experiments, it is important to analyze them vis-a-vis domain expert reasoning. An interesting piece of research proposes a technique called FeaturesRank \cite{varde2007learning} to learn the relative importance of various features in scientific images, including those at nanoscale levels. Considering a more generic perspective, AI researchers propose an approach called YOLO (You Only Look Once) for fast real-time object detection, addressing 9000+ object categories \cite{Redmon2016YOLO9000BF}. Furthermore, the fusion of vision and language is proposed in the context of object recognition \cite{shiang2017spatialhelpsvision}, presented at AAAI. These and other studies in object identification and image classification are highly significant in scientific domains because they help in efficient analysis of a large number of images from lab experiments and industrial processes, thereby aiding scientists in decision-making. 

Moving forward, commonsense knowledge can be crucial in various AI-related applications as elaborated in the works of many researchers \cite{tandon2018commonsense}, \cite{davis2015commonsense}, \cite{ilievski2021dimensions}, \cite{razniewski2021information}, \cite{kaiser2014extracting}, \cite{Nguyen2021WWW}. Some of these pertain to science and engineering domains. We humans have inherent common sense that helps us distinguish objects within images, and associate everyday facts with scientific concepts, in order to assist our natural decision-making even in situations encountered for the first time. Since huge amounts of prior training data may not always be available, it is advisable to complement the learning of any AI-based systems with knowledge bases incorporating commonsense knowledge. State-of-the-art systems such as TupleKB \cite{mishra2017domain} and Arc \cite{clark2018think} provide commonsense knowledge bases, and aid in commonsense reasoning pertinent to scientific facts; hence they can be used in conjunction with various AI systems. 

Considering various such related work, our research in this paper makes a contribution towards AI and Materials Science with a view to aiding sustainability and helping in the overall vision of saving the planet. Our approaches that aim to use recycled \& waste materials in agile sustainable manufacturing, relying on machine learning techniques for predictive analysis, make impacts on green and clean manufacturing, thus being in line with sustainability. 

\section{Conclusions and Open Issues}
In conclusion, we re-emphasize that our prior work on AutoDomainMine for computational estimation motivated this research, encouraging us to pursue further studies. Related work by many researchers in AI, Materials Science and sustainability provided further inspiration to conduct studies on agile sustainable manufacturing. 
In this paper, we aim to reuse recycled and waste materials from forward operating bases, in order to generate valuable products (for use in weaponry to aid defense operations), thereby making positive impacts on environmental sustainability. We explore machine learning techniques such as ANN, RF and CNN using real data from lab experiments in order to conduct predictive analysis to guide decision support in industrial manufacturing processes. Our ANN results incur accuracy around 90\%, high efficiency, and low costs (lab experiment costs only). CNN and RF also provide accuracy in a very high range, aiding deeper analysis on a finer-level, and better comprehensibility for explainable AI, respectively.  

Future challenges include investigating computer vision models such as VGG, ResNet etc. to potentially enhance accuracy, efficiency and robustness. We aim to build tools using machine learning for decision support in agile manufacturing. We also plan to assess the precise impacts on sustainability by proposing and deploying adequate metrics based on the use of recycled materials, and the effectiveness of developed products. In general, this work treads the bridge connecting AI and Materials Science, while also making positive impacts on environmental sustainability. 

\section*{Acknowledgments}
A. Varde acknowledges NSF grants ``MRI: Acquisition of a High-Performance GPU Cluster for Research and Education'' Award Number 2018575, and ``MRI: Acquisition of a Multimodal Collaborative Robot System (MCROS) to Support Cross-Disciplinary Human-Centered Research and Education at Montclair State University'', Award Number 2117308. She is a visiting researcher in the group of Dr. Gerhard Weikum at Max Planck Institute for Informatics, Saarbrücken, Germany, ongoing from her sabbatical.

\bibliography{references}
\bibliographystyle{aaai}

\end{document}